# Patients' Severity States Classification based on Electronic Health Record (EHR) Data using Multiple Machine Learning and Deep Learning Approaches

A. N. M. Sajedul Alam[1], Rimi Reza[1], Asir Abrar[1], Tanvir Ahmed[1], Salsabil Ahmed[1], Shihab Sharar[1], Annajiat Alim Rasel[1]

[1] Department of Computer Science and Engineering, School of Data and Sciences (SDS), BRAC University, 66 Mohakhali, Dhaka 1212, Bangladesh
{a.n.m.sajedul.alam, rimi.reza, asir.abrar, tanvir.ahmed8, salsabil.ahmed, shihab.sharar}@g.bracu.ac.bd, annajiat@bracu.ac.bd

**Abstract.** This research presents an examination of categorizing the severity states of patients based on their electronic health records during a certain time range using multiple machine learning and deep learning approaches. The suggested method uses an EHR dataset collected from an open-source platform to categorize severity. Some tools were used in this research, such as openRefine was used to pre-process, RapidMiner was used for implementing three algorithms (Fast Large Margin, Generalized Linear Model, Multi-layer Feed-forward Neural Network) and Tableau was used to visualize the data, for implementation of algorithms we used Google Colab. Here we implemented several supervised and unsupervised algorithms along with semi-supervised and deep learning algorithms. The experimental results reveal that hyperparameter-tuned Random Forest outperformed all the other supervised machine learning algorithms with 76% accuracy as well as Generalized Linear algorithm achieved the highest precision score 78%, whereas the hyperparameter-tuned Hierarchical Clustering with 86% precision score and Gaussian Mixture Model with 61% accuracy outperformed other unsupervised approaches. Dimensionality Reduction improved results a lot for most unsupervised techniques. For implementing Deep Learning we employed a feed-forward neural network (multi-layer) and the Fast Large Margin approach for semi-supervised learning. The Fast Large Margin performed really well with a recall score of 84% and an F1 score of 78%. Finally, the Multi-layer Feed-forward Neural Network performed admirably with 75% accuracy, 75% precision, 87% recall, 81% F1 score.

# 1   Introduction

In this current scientific era, many machine learning approaches are employed to improve patient outcomes by harnessing the rising amount of health data provided by the IoMT(Internet of medical things). Machine learning has been used in three major categories in recent years: computer-aided diagnosis, natural language processing of clinical publications, as well as genetic information. Artificial intelligence will almost certainly have a substantial influence on the health business in the future, helping both patients and professionals.

In an emergency, obtaining reliable data rapidly is critical to diagnosing and addressing medical concerns. Artificial intelligence may be used by all types of doctors, nurses, and medical personnel to employ real-time, accurate figures to accelerate up and enhance crucial treatment choices. Improved preventative actions, cost savings, and patient wait times may all be achieved by generating more rapid and realistic results.

An Electronic Medical Record (EHR) Data means a digital record of a condition and its treatments that a doctor maintains track of overtime time. It might include all of the key administrative medical information relevant to that patient's care with a specific physician. The EHR simplifies information availability and also has the power and potential to enhance clinical workflow. The availability and quality of this knowledge will enable physicians to make informed decisions while also providing effective therapy.

The level of organ system malfunction or metabolic hypotension for a patient is characterized as the illness severity.  It categorizes medical conditions as mild, moderate, significant, or extreme. This categorization is intended to serve as a foundation for monitoring hospital resource utilization or developing patient care standards.

With a vast technological advancement, still, in many hospitals, people are dying in need of doctors and lack the number of staff to take care of them. With the growing population of the earth, this problem is growing day by day globally. If there were sufficient automated technical medical support based on IoMT as well as AI, many lives could have been saved.

At the beginning of our research we found that in many countries, hospitals are still not that much using any Advanced Artificial Intelligence or the Internet of Medical

Things to make things easier and faster. W have shown in our research that it is possible to categorize the severity of illness by using machine learning and deep learning approaches, we hope at least some other researchers could have grown interested in solving this problem with better solutions. Already many researchers have shown interest in this and worked on this, we got inspired by their works.

This study's overarching purpose is to improve on previous work [6] and to try out new approaches to this specific clinical data and gain relevant insights from it in order to deliver new resources and innovations to the global community for a faster Artificial Intelligence and Internet of medical things-powered clinical decision assistance to patients in critical situations. Which can be used by both patients and physicians.

## 2  Literature Review

Various ML techniques have been proved in studies to detect a wide range of medical diagnoses and curable disorders. Several researchers have lately attempted to apply this strategy to the major COVID-19 quandary. Cough sound classification is used from one of the latest COVID-19 crisis-prompted research [1] to detect lung disorders. They employed SVM Classifier, Convolution Neural Network as well as Artificial Neural Network,  for solving the classification problem.

Another Deep Learning-based model [2] was created to predict the deterioration of patients' health conditions so that appropriate action can begin as soon as possible. RNN, as well as LSTM, were utilized in this research study as Deep Learning Techniques. Finally, with a precision of much more than 80%, this system was able to reliably predict the severity.

Predictions based on digital health record data are nowadays becoming increasingly common. Many forms of health-related forecasts and detections are being implemented using ML methods. A model [3] has been built to identify mental diseases by including six classification algorithms, which are a tuned neural network, XGBoost, Random Forest, SVM Classifiers, and K-nearest Neighbor. Using Deep Learning networks on EHR, a model for detecting and visualizing bleeding episodes [4] has been suggested.

A secondary usage of electronic health record data, as Miotto et al. [5] have out, has the capacity to boost clinical research and clinical decision-making. Predictive modeling using EHRs is not commonly used due to the difficulties in gathering and expressing patient data. For constructing overall patient models from EHR data for use in clinical predictive modeling they created a unique unsupervised deep feature learning method. Their findings suggest that deep learning in EHRs can provide patient models that support healthy prediction, and that deep learning could be used to complement clinical decision systems.

The COVID era implies an urgent requirement to identify the seriousness levels of patients given the fact that large numbers of people are being infected on a regular basis. Sadikin et al. [6] presented a patient treatment categorization model based on several machine learning algorithms. The desired dataset, EHR, was collected from an Indonesian clinic. When contrasted to other classifiers inside the tests, hyperparameter-tuned XGBoost obtained the highest accuracy level of 0.7579.

EHRs, according to Liu, Ziyi, et al. [49], include significant information about a disease and its treatment, which often contains both structured and unstructured data. A lot of research has been conducted to extract important information from structured data such as illness codes, test findings, and therapies. However, relying just on structured data may be insufficient for effectively recording patients' whole data, and such data may occasionally contain inaccurate entries. With recent advances in Deep Learning as well as Machine Learning methodologies, an increasing number of academics are striving to obtain more accurate results by integrating unstructured unlimited services as well.

Scores from the National Institutes of Health Stroke Scale are regularly used to quantify stroke severity, which is a major predictor of patient outcomes, according to Kogan, Emily, et al. [50]. Because these evaluations are frequently supplied in physician reports as loose language, organized real evidence collections rarely include the severity. The study's goal was to apply deep learning methods to forecast NIHSS scores for all persons with acute stroke using electronic medical record data from several institutions.

## 3 Proposed Methodology

This study was conducted as a follow-up to the publication [6] where they found that XGBoost (hyperparameter-tuned) outperformed other supervised methods and they did not try out any other approaches for instance deep learning, semi-supervised and unsupervised approaches, here in this research the goal of this work was to improve the previous work [6] and also to apply different approaches such as Unsupervised, Semi-Supervised, Deep Learning models along with Supervised models to a specific Electronic Health Record dataset which was also used in [6] and evaluate their effectiveness in classifying patients' sickness severity levels. SVC, XGBoost, RandomForest, NaiveBayes, LightGBM, K-nearest Neighbor, Decision-Tree, CatBoost, AdaBoost, Logistic Regression, and Generalized Linear Model were the supervised models employed in this study. We used K-means clustering, Spectral

Clustering, Gaussian Mixture Model, and Hierarchical Clustering for unsupervised techniques. The Semi-supervised methodology was built using a Fast Large Margin method. Furthermore, in the case of applying deep learning, we constructed a Multi-layer feed-forward neural net.

## 3.1 Dataset

The dataset was gathered through the open-source dataset platform 'Mendeley Data' [40]. The information used in the study was Electronic Patient Record Data Predicting obtained from a hospital in Indonesia which was private, according to the report. Contains the patient's laboratory report data, which were used to decide whether the person is in or out of care, as well as the patient's future course of treatment.

The dataset used in this study includes both laboratory reports and demographic data, as shown in **Table 1**. To describe the table, the first column, 'Name,' contains all of the attribute names, and the second column, 'Data Type,' indicates the type of data held, which can be either Numerical-Continuous or Categorical-Nominal. The third column, titled 'Value Sample,' displays a single sample value for that particular feature. As an example, the sample value for the attribute 'Age' is 12 or it can be any age in between 1 to 99 and 'Sex' attribute's values are binary, such as either 'Male' or 'Female', also the output column name is 'Severity Level', outputs are also binary such as either 'Mild' or 'Severe' which indicates towards the severity of illness of a patient. The final column, 'Description,' contains a brief description of the characteristic.

**Table 1.** Result Analysis Of Supervised Classification Algorithms.

| Name | Data Type | 1-Value Sample | Description |
|---|---|---|---|
| MCHC | Numerical-Continuous | 33.6 | Patient's Lab report of MCHC |
| SEVERITY LEVEL | Categorical-Nominal | Mild, Severe | Severity level of Patient's illness |
| HEMOGLOBINS | Numerical-Continuous | 11.8 | Patient's Lab report of Hemoglobins |
| AGE | Numerical-Continuous | 12 | Age of the patient |
| THROMBOCYTE | Numerical-Continuous | 310 | Patient's Lab report of Thrombocyte |
| SEX | Categorical-Nominal | Binary | Gender of the patient |
| MCH | Numerical-Continuous | 25.4 | Patient's Lab report of MCH |
| LEUCOCYTE | Numerical-Continuous | 6.3 | Patient's Lab report of Leucocyte |
| MCV | Numerical-Continuous | 75.5 | Patient's Lab report of MCV |
| ERYTHROCYTE | Numerical-Continuous | 4.65 | Patient's Lab report of Erythrocyte |
| HEMATOCRIT | Numerical-Continuous | 35.1 | Patient's Lab report of Hematocrit |

| | HAEMATOCRIT | HAEMOGLOBSevereS | ERYTHROCYTE | LEUCOCYTE | THROMBOCYTE | MCH | MCHC | MCV | AGE | SEX | SEVERITY LEVEL |
|---|---|---|---|---|---|---|---|---|---|---|---|
| 2 | 35.1 | 11.8 | 4.65 | 6.3 | 310 | 25.4 | 33.6 | 75.5 | 1 | F | Mild |
| 3 | 43.5 | 14.8 | 5.39 | 12.7 | 334 | 27.5 | 34 | 80.7 | 1 | F | Mild |
| 4 | 33.5 | 11.3 | 4.74 | 13.2 | 305 | 23.8 | 33.7 | 70.7 | 1 | F | Mild |
| 5 | 39.1 | 13.7 | 4.98 | 10.5 | 366 | 27.5 | 35 | 78.5 | 1 | F | Mild |
| 6 | 30.9 | 9.9 | 4.23 | 22.1 | 333 | 23.4 | 32 | 73 | 1 | M | Mild |
| 7 | 34.3 | 11.6 | 4.53 | 6.6 | 185 | 25.6 | 33.8 | 75.7 | 1 | M | Mild |
| 8 | 31.1 | 8.7 | 5.06 | 11.1 | 416 | 17.2 | 28 | 61.5 | 1 | F | Mild |
| 9 | 40.3 | 13.3 | 4.73 | 8.1 | 257 | 28.1 | 33 | 85.2 | 1 | F | Mild |
| 10 | 33.6 | 11.5 | 4.54 | 11.4 | 262 | 25.3 | 34.2 | 74 | 1 | F | Mild |
| 11 | 35.4 | 11.4 | 4.8 | 2.6 | 183 | 23.8 | 32.2 | 73.8 | 1 | F | Mild |
| 12 | 33.7 | 11.5 | 4.57 | 13.2 | 322 | 25.2 | 34.1 | 73.7 | 1 | M | Mild |
| 13 | 54 | 16.6 | 7.61 | 10 | 88 | 21.8 | 30.7 | 71 | 1 | F | Severe |
| 14 | 31.7 | 10.4 | 4.91 | 9.7 | 348 | 21.2 | 32.8 | 64.6 | 1 | M | Severe |
| 15 | 35.3 | 11.9 | 4.4 | 5.8 | 205 | 27 | 33.7 | 80.2 | 1 | M | Mild |
| 16 | 34.5 | 9.8 | 5.75 | 15.4 | 548 | 17 | 28.4 | 60 | 1 | M | Mild |
| 17 | 34 | 10.3 | 5.27 | 16.2 | 572 | 19.5 | 30.3 | 64.5 | 1 | M | Mild |
| 18 | 35 | 11.6 | 4.58 | 7.4 | 154 | 25.3 | 33.1 | 76.4 | 1 | F | Mild |
| 19 | 51.3 | 15.7 | 7.24 | 4.8 | 129 | 21.7 | 30.6 | 70.9 | 1 | F | Mild |
| 20 | 31.3 | 10.8 | 4.02 | 7.9 | 250 | 26.9 | 34.5 | 77.9 | 1 | F | Mild |
| 21 | 36.8 | 12.9 | 4.67 | 5.7 | 235 | 27.6 | 35.1 | 78.8 | 1 | F | Mild |

**Fig. 1.** First 20 samples of our dataset. In **Fig. 1** the first 20 samples of our dataset were shown to understand what kind of data we are working with and in which structure the dataset is in.

**Table 2.** Count of unique values for each attributes in the dataset

| Attributes | Count of unique values |
|---|---|
| Hematocrit | 326 |
| Hemoglobins | 128 |
| Erythrocyte | 433 |
| Leucocyte | 276 |
| Thrombocyte | 554 |
| MCH | 189 |
| MCHC | 105 |
| MCV | 406 |
| Age | 95 |
| Sex | 2 |
| Severity Level | 2 |

From **Table 2** we can see that the highest number of unique values belong to Thrombocyte and the second-highest number of unique values belong to Erythrocyte. Also, MCV has the third-highest number of unique values.

## 3.2 Data Visualization

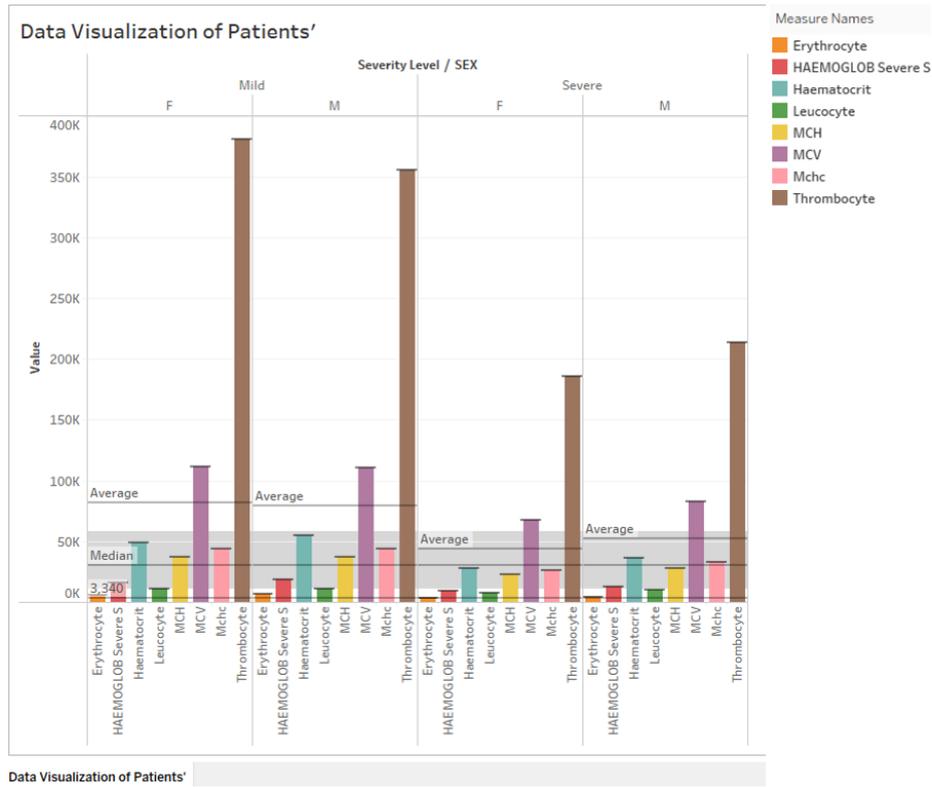

**Fig. 2.** Data visualization of patients'. In **Fig. 2** we tried to show the visualization of patients' overall data we have along with the average, median levels of data for each feature. This visualization was divided into two major categories which are Mild and Severe, also these two were also divided into two categories by genders, Male and Female for presenting data in a more organized manner.

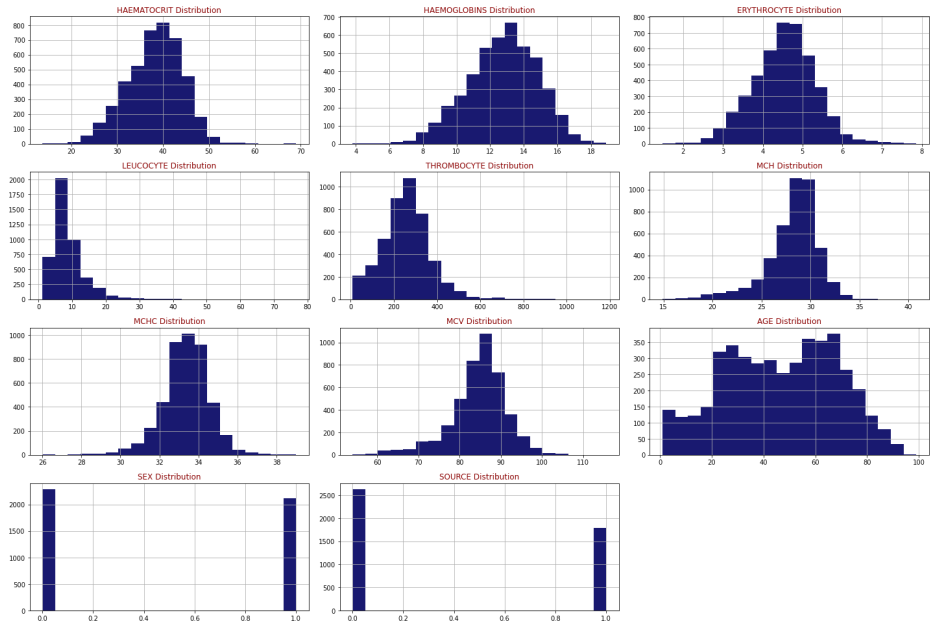

**Fig. 3.** Distribution of each feature. In **Fig. 3**, it was shown how each attribute got distributed in our dataset by using histograms. Here 'Source' means the severity of illness. In these histograms, we had used 20 bins for visualizing distributions and skewnesses of each feature in the dataset. From **Fig. 3** we can see that the distribution of 'MCV', 'MCHC' features are right-skewed or we can also say that they are positively skewed when the distribution of the 'AGE' feature is left-skewed or negatively skewed. The distribution of the 'Sex' and 'SOURCE' features is **Bernoulli**. On the other hand, the distribution of 'MCV', 'MCHC', 'AGE', 'MCH', 'THROMBOCYTE', 'LEUCOCYTE', 'ERYTHROCYTE', 'HEMOGLOBINS' and 'HEMATOCRIT' is **Binomial**.

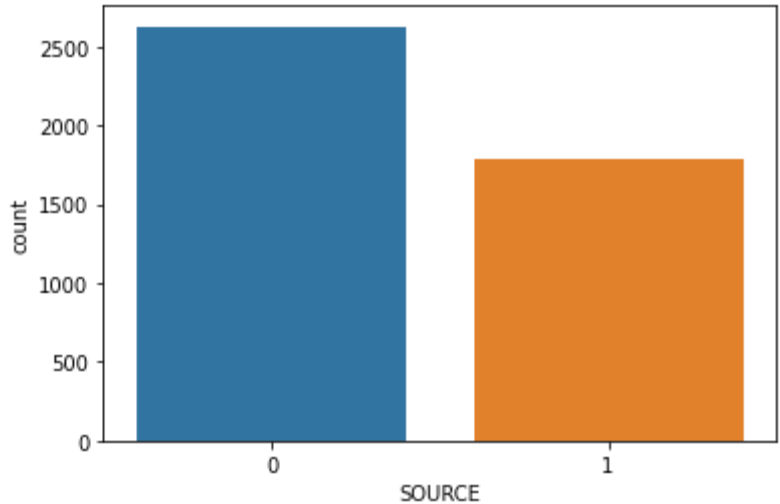

**Fig. 4.** Countplot for the attribute SOURCE which represents the severity states of illness. In **Fig. 4**, we visualized the counterplot of the SOURCE attribute, which is basically the illness severity states of patients. Here, **0** means out or out-care, which means a mild state of illness, and **1** means in or in-care, which means a severe state of illness. From **Fig. 4** we got a really important insight into this dataset, which is that here more patients are in out-care or in a mild state of their illness and fewer patients are in in-care or severe state of their illness. To be more specific with calculations, there are nearly **1550** patients in in-care or severe state of their illness and more than **2500** patients in an out-care or mild state of their illness.

### 3.3 Data Cleaning and Preprocessing

To minimize biased results when classifying using openRefine, the data was scaled, cleaned, and standardized before applying supervised machine learning classifiers. Dimension Reduction (DR) was done for unsupervised learning techniques by looking at the heatmap [15] that showed the correlation matrix [23] between distinct features of the dataset. When clustering, a high correlation among independent features might lead to biases in the findings. As a result, strongly linked features were eliminated from the study. The correlation chart shows that 'HAEMATOCRIT', 'HAEMOGLOBINS', 'ERYTHROCYTE', 'MCH', and 'MCHC' are strongly linked characteristics, therefore they were excluded. In addition, the accuracy of this condensed dataset was compared to prior dataset results.

### 3.3.1 Feature Selection

While creating a predictive model, feature selection is really important. It is a method of minimizing the number of independent variables. We utilized the dimension reduction approach to identify the most essential characteristics from all of the features in this case.

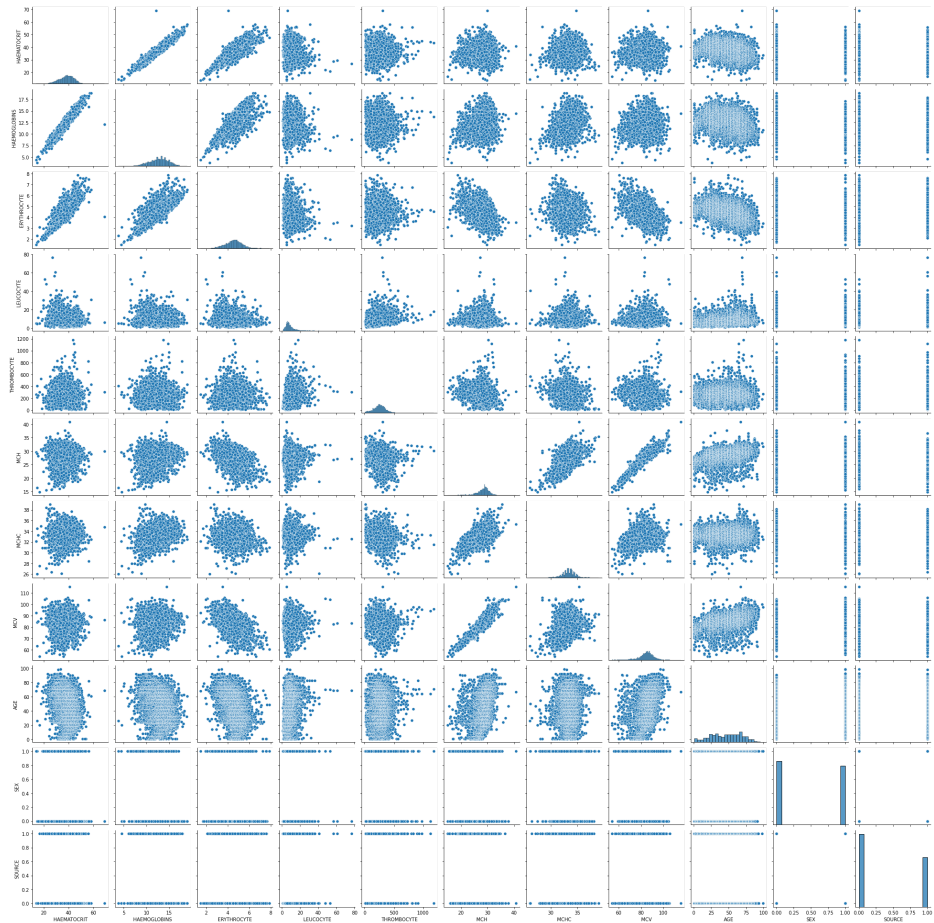

**Fig. 5.** Pair plot of each feature in the dataset. In **Fig. 5**, we visualized a pair plot of each feature in the whole dataset to show the relation between features to understand and explain the interrelationship of attributes which helped us choose important features as well. Additionally, it gave direction towards what kind of algorithms should be used on this dataset.

### 3.3.2 Dimension Reduction

With a large set of measurements in the feature subspace, the area of that space can be quite large, and the endpoints (rows of information) that we have in that space typically represent a small and non-representative sample. Methods for minimizing the number of independent variables in the training dataset are referred to as dimensionality reduction. The 'Feature Selection Method' [5] is one dimension reduction approach. Dimension Reduction was used in unsupervised learning methods.

**Correlation Matrix:** A strong correlation between dependent and independent variables is desirable, whereas a strong correlation between two independent variables is undesirable [2]. This notion is used for feature extraction to apply dimension reduction.

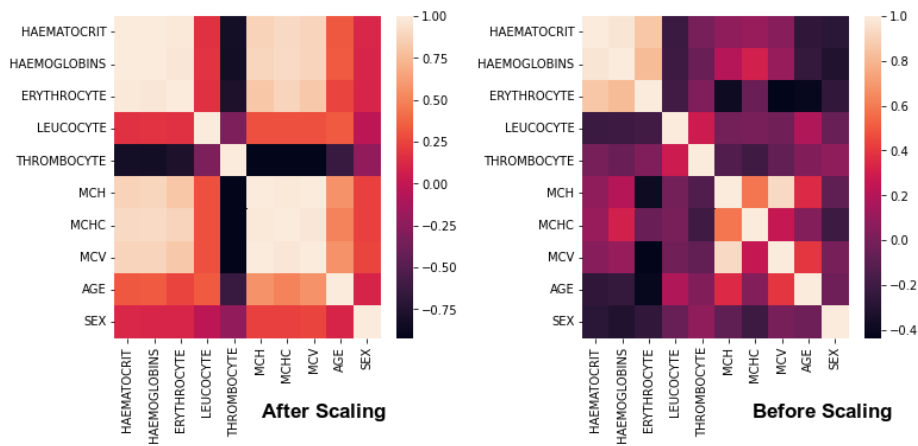

**Fig. 6.** Before and after scaling, the heatmap of all independent variables. This is the heatmap representation of the correlation matrix of independent variables before and after scaling the data in **Fig. 6**. From **Fig. 6**, we can see that correlation of Erythrocyte and Hematocrit, Erythrocyte and Hemoglobin are really higher than the correlation of other features. On the other hand, the correlation between 'MCH' and 'MCV', 'MCH' and 'MCHC' is also significant. For that reason, we can stand on this hypothesis that if we pick 'ERYTHROCYTE', 'HEMATOCRIT', 'HEMOGLOBINS', 'MCH', 'MCV', and 'MCHC' as our main features, the models can give us better results.

## 3.4 Approach of Unsupervised Learning of Machine

The unsupervised learning methodology, also known as the unsupervised machine learning model, employs machine learning algorithms to analyze and cluster unlabeled data. Without the need for human intervention, these algorithms uncover underlying patterns or data categories. It is the best choice for data discovery, cross-selling strategies, consumer segmentation, and image identification because of its capacity to identify similarities and contrasts in information. Individual health records were used to forecast the severity of the patients, therefore clustering algorithms were used to aggregate data. The processes for each clustering model will be discussed below.

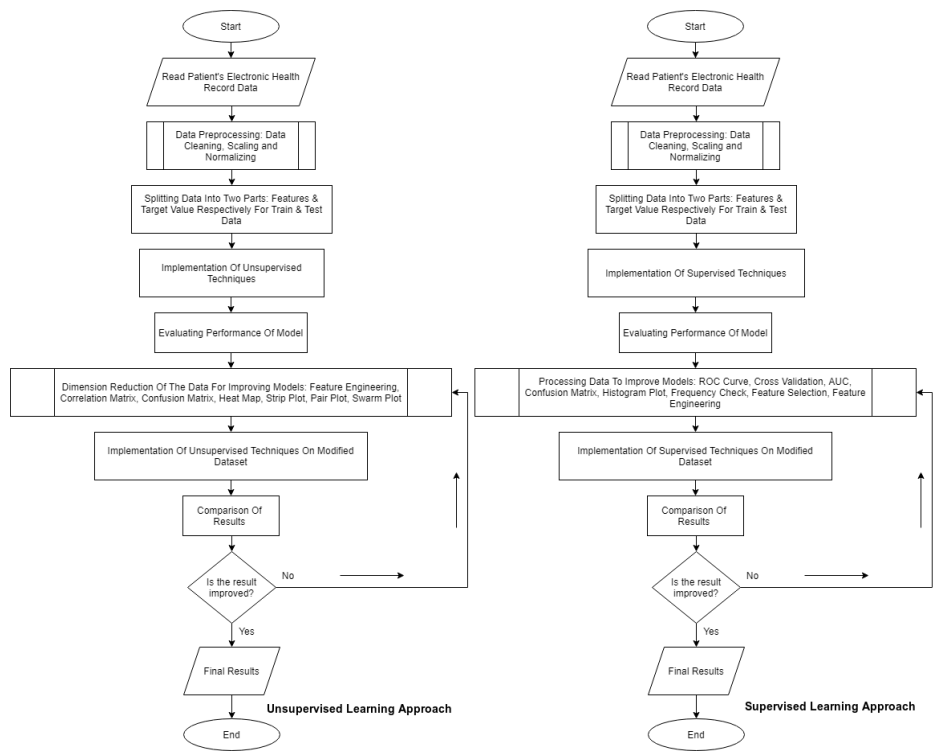

**Fig. 7.** Workflow of unsupervised and supervised learning approaches of the machine. In **Fig. 7** we can see that the unsupervised learning approach starts with reading the Patient's Electronic Health Data. After reading, data is processed, cleaned, scaled, and normalized. And after this process, data is divided into two parts. Then the unsupervised techniques are implemented. After implementing unsupervised techniques, model performance is evaluated. And after this process, the process of improving the result is conducted. Here dimension reduction for feature engineering, correlation matrix, confusion matrix, heat map, strip plot, pair plot, and swarm plot is done. After this, the implementation of unsupervised techniques and comparison is done. If the results are improved, the final result is shown. After getting the final result the process ends. As we can see from **Fig. 7** the supervised learning technique begins with examining the patient's electronic health data. Data is processed, cleaned, scaled, and normalized after reading. Following this, the data is separated into two pieces. The monitored approaches are then put into action. Model performance is assessed following the use of supervised approaches. Following this, the process of enhancing the result is carried out by improving the ROC curve, cross-validation, AUC, confusion matrix, histogram plot, frequency check, feature selection, and feature engineering. Following that, supervised procedures are implemented and compared. The ultimate outcome is displayed if the results are improved. The process concludes once the final result is obtained.

### 3.4.1 Unsupervised Algorithms

I. **K-means Clustering:** This technique [41] is basically a type of unsupervised learning that may be used with unlabeled data means data without defined groups or categories to be specific. The goal of this approach is to find groups in data, where K is the number of categorizations or groups. Iteratively, the method allocates each data point to one of the K groups or categories based on the qualities supplied. Data points are clustered collectively based on how similar their characteristics are. [5]

II. **Elbow diagram:** In K-means clustering, the elbow method was implemented on the dataset to justify how many clusters should be taken in the count to proceed for getting highly efficient clustering results [18], [45].

III. **Hierarchical Clustering:** Prior to using Hierarchical Clustering [43], the data should be normalized such that each variable has the same scale. The significance of this is that the scales of the variables aren't the same, the model may be skewed toward variables of greater size.

IV. **Dendrogram:** A dendrogram is a type of tree that represents hierarchical grouping relationships between similar data sets. They are commonly used in biology to show gene pools or patterns, but they can represent any type of clustered data [19]. The dendrogram helps to decide the number of clusters for solving this particular problem.

- V. **Spectral Clustering:** Spectral clustering [20] is a method used in EDA to divide huge multidimensional datasets into tiny groupings with related data in rarer dimensions. The primary objective is to group all unstructured data pieces into several groups based on similarities. The dataset was converted into a particular numeric value rather than a random collection of values from various ranges. Furthermore, the data were scaled and normalized. It was then translated from a NumPy array to a pandas dataset. Finally, PCA analysis was used to minimize the dimensionality of the data [13].
- VI. **PCA (Principal component analysis):** A machine learning strategy that uses unsupervised learning to minimize dimensionality. This research includes Spectral Clustering based on the PCA technique. After PCA analysis [24], the spectral clustering model is trained on that dataset.
- VII. **Gaussian Mixture Model:** It is assumed that there exist a finite amount of Gaussian distributions, which are represented by a cluster. As a result, with such a Gaussian Distribution, measured values from the single distribution are much more prone to be clustered together [44].

## 3.5 Approach of Supervised Learning of Machine

The technique of training a mechanism that changes the input into output based on instances of input-output pairs is known as supervised learning. It creates a training case. In a supervised approach, there should be always a set of predefined examples.

### 3.5.1 Supervised Algorithms

- I. **Support Vector Machine (SVM) classifier:** This approach can classify both linear and non-linear data. It starts by mapping data to something like an n-dimensional domain [20], where n represents the count of features. The hyperplane that separates the data into classes is then calculated [32], [24].
- II. **Decision Tree (DT):** This is basically a supervised non-parametric learning model for classification and regression problems [36, 35]. This tree-structured classifier's internal nodes contain dataset properties, branches show the forecasting models, and each leaf node offers the conclusion [42].
- III. **XGBoost:** This is a machine learning strategy that can handle missing data without requiring imputation preprocessing and is widely utilized for classification problems [39].
- IV. **AdaBoost:** This technique allows weak classifiers to improve their performance by adaptively altering the weak learning cycle [28]. It performs well on well-balanced datasets but suffers when there is noise [31].

V. **LightGBM:** This is yet another gradient boosting method that makes use of tree-based learning methods. It outperforms XGBoost [24] in terms of computation speed and memory utilization.
VI. **KNN:** It is known as a lazy training algorithm because, instead of learning instantly from the training dataset, it keeps the data and performs actions on it until it time comes to categorize it [22], [1].
VII. **Random forest:** This is a tree prediction class in which the values of random variables created at random with almost the same distribution any as well as all trees inside the forest are used to know the outcome of individual trees [32], [3], [6], [30].
VIII. **CatBoost:** This is a publicly available as well as free Gradient Boosted Decision Tree (GBDT) method for Supervised Machine Learning. This strategy, which is based on Decision Trees, performs exceptionally well on Machine Learning problems that need heterogeneous as well as categorical data [15].
IX. **Naive Bayes Classification:** This method is a simple as well as effective Classification technique that aids in the development of quick machine learning able to produce correct predictions. It is a forecasting method that makes estimation based on an item's probability [52].
X. **Logistic Regression:** This technique is a statistical approach that estimates the chance of a data point using supervised learning. Because the objective or moderating factors have a dichotomous structure, there are only two viable categories [51].
XI. **Generalized Linear Model:** This is just a sophisticated statistical modeling approach developed in 1972 by both John Nelder and Robert Wedderburn. It is a catch-all phrase for a variety of alternative models that enable the suitable y to provide a cause of error besides a normally distributed model [53].

## 3.6 Approach of Semi-Supervised Learning of Machine

This is a mixture of unsupervised as well as supervised approaches. Building a machine learning technique on labeled data, with each recording including outcome information, is a common supervised machine learning strategy.

### 3.6.1 Semi-Supervised Algorithm

**Fast Large Margin:** Chang et al. [56] suggested a fast margin learner focused on a support vector linear learning technique for the fast big margin approach. Additionally, this semi-supervised approach can work with a huge dataset having a lot of attributes and instances easily.

## 3.7 Deep Learning Approach

Deep neural networks may attain cutting-edge performance, often outperforming humans. Networks are built to utilize a vast quantity of class labels and multi-layered neural net designs.

### 3.7.1 Deep Learning Algorithm

**Feed-Forward Artificial Neural Network (Multi-Layer):** The network might include numerous hidden layers composed of neurons with rectifier, tanh as well as max out activation functions.[57] Back-propagation gradient descent (stochastic) is used to train this model of neural network.

## 4 Result Analysis

This research study employs a variety of evaluation measures. Recall, Precision, F1 as well as Accuracy scores had been observed in all learning strategies, whereas additional clustering assessment approaches such as silhouette score, BIC score, and the log-likelihood score had been observed in the unsupervised learning approach. These metrics vary depending on the number of dimensions reduced and other changes that occurred when the model hyper-parameters were changed.

From the mentioned below (**1**), (**2**)**,** and (**3**) number equations, **TP** means True Positive, which indicates that the actual positive values and predicted positive values got matched. **FP** means False Positive, which indicates original positive values as well as predicted positive values did not get matched. Additionally, **TN** means True Negative, which tells us that predicted negative values, as well as original negative values, are the same. On the other hand, **FN** means False Negative, which shows that the predicted negative values and actual values are not exact.[61]

I. **Precision:** Precision is equal to the number of Positive Instances divided by the total of True Positives as well as False Positives.

$$TP / (TP + FP) = Precision. \tag{1}$$

II. **Recall:** The recall's True Positives are divided by the recall's True Positives as well as False Negatives.

$$TP / (TP + FN) = Recall. \tag{2}$$

III. **Accuracy:** In multilabel classification, this method computes subset accuracy: the subset of labels forecasted for a sample must completely match the corresponding true label value [21].

$$(TP + TN) / (TP + TN + FP + FN) = Accuracy. \tag{3}$$

IV. **F1 score:** The F1 score represents the balance between accuracy and recall [10], [26].

$$2 * ((Precision * Recall) / (Precision + Recall)) = 2 * TP / (2\ TP + FP + FN) = F1. \tag{4}$$

V. **Silhouette score:** The term "silhouette" refers to a strategy for examining and confirming data set consistency. The approach provides a simple graphical representation of how well each object was categorized. The silhouette value reflects how close an item is to its group in comparison to other clusters [47], [29].

$$(B_p - A_p) / Max(B_p, A_p) = Silhouette. \tag{5}$$

Here, in equation (**5**), **p** is a specific data point and for that specific data point, silhouette score can be calculated using this equation. **A**, **B** is respectively, inter-cluster and intra-cluster distance.[59]

VI. **BIC score:** This is a model scoring as well as a selection technique which is known as Bayesian Information Criterion. It is named after the subject of study in which it arose: probabilities and as well as inference within Bayes theory. As the model's complexity increases, so does the BIC value, and as the probability increases, so does the BIC value. As a result, a smaller number is preferable.

$$r * \ln(q) - 2 * \ln(M) = BIC . \tag{6}$$

Here in equation (**6**) **M** is the likelihood function's[62] maximum value, **r** is the model's predicted number of parameters, **q** is the sample size.[60]

VII. **Log-likelihood score:** A log-likelihood value is the measure of each model's quality of fit. The better the model, the higher the value. It is important to understand that the logarithm probability might range from Inf and + Inf. As a result, a statement based only on the value cannot be made. Only log probability values may be compared between models [16].

Table 3. K-Means Clustering Evaluation Metrics

| Dimension Reduction | Precision | Recall | Accuracy | F1 Score | Silhouette Score |
|---|---|---|---|---|---|
| Before | 0.3036 | 0.3498 | 0.3251 | 0.4127 | 0.3074 |
| After | 0.4922 | 0.6502 | 0.5602 | 0.5873 | 0.4672 |

Table 4. Hierarchical clustering evaluation before Dimension Reduction

| Hyper Parameter: linkage | Precision | Recall | Accuracy | F1 Score | Silhouette Score |
|---|---|---|---|---|---|
| Ward | 0.3296 | 0.6586 | 0.3203 | 0.4393 | 0.6296 |
| Complete | 0.3569 | 0.7926 | 0.3386 | 0.4922 | 0.6297 |
| Average | 0.3726 | 0.8587 | 0.3583 | 0.5197 | 0.6781 |

**Table 5.** Hierarchical clustering evaluation after Dimension Reduction

| Hyper Parameter: linkage | Precision | Recall | Accuracy | F1 Score | Silhouette Score |
|---|---|---|---|---|---|
| Ward | 0.3168 | 0.6003 | 0.3150 | 0.4147 | 0.5660 |
| Complete | 0.3569 | 0.7926 | 0.3386 | 0.4922 | 0.6677 |
| Average | 0.3739 | 0.8699 | 0.3585 | 0.5231 | 0.7004 |

**Table 6.** Gaussian Mixture Model before dimension reduction

| Hyper Parameter | Precision | Recall | Accuracy | F1 Score |
|---|---|---|---|---|
| Default | 0.6638 | 0.4361 | 0.6826 | 0.5263 |
| Changed | 0.3104 | 0.5639 | 0.3173 | 0.4004 |

**Table 7.** Gaussian Mixture Model after dimension reduction

| Hyper Parameter | Precision | Recall | Accuracy | F1 Score |
|---|---|---|---|---|
| Default | 0.3107 | 0.5779 | 0.3109 | 0.4041 |
| Changed | 0.3107 | 0.5779 | 0.3109 | 0.4041 |

**Table 8.** Spectral clustering Analysis

| Hyper Parameter | Dimension Reduction | BIC Score | Log-Likelihood Score |
|---|---|---|---|
| Default | before | -299093.3968 | 34.0200 |
| Changed | before | -299093.4942 | 34.0200 |
| Default | after | -299093.3969 | 12.4517 |
| Changed | after | -109529.8661 | 12.4517 |

Here from **Table 3,** it was observed that after dimension reduction K-means clustering algorithm performed better according to every performance metric on the table. From **Table 4** and **5**, it was easily observed that, without dimension reduction, the linkage average hyperparameter was performing best and the second best was the linkage complete hyperparameter according to the Recall score. On the other hand, after dimension reduction, for both of the linkage hyperparameters, the Recall score was improved. Additionally, according to Silhouette, F1, Accuracy, and Precision scores, the scores for these two linkages respectively complete and average were improved for all scores. **Table 6** and **7** describes the evaluation metrics for Gaussian Mixture models. Prior to dimension reduction, the implementation of hyperparameters improved just the recall score. Furthermore, adding hyperparameters had no effect on any scores after dimension reduction. **Table 8** refers to the spectral clustering analysis based on BIC score and log-likelihood score. There were no changes in any scores for introducing hyperparameters without dimension reduction. Although, without applying hyperparameters, the Log-Likelihood score deteriorated after dimension reduction, and with hyperparameters modified after dimension reduction, the BIC score dropped.

This research study employed a variety of evaluation measures. Precision, recall, accuracy, and F1 score had been seen in the supervised, unsupervised, semi-supervised machine learning and deep learning strategy. These metrics vary depending on the number of dimensions reduced and other changes that occurred when the model hyperparameters were changed. We had already discussed this in the **3.3 Data Cleaning and Preprocessing** segment in our paper.

Table 9. Description of the legends used in unsupervised results.

| Legends | Descriptions |
| --- | --- |
| K-Means1 | K-Means Clustering Before Dimension Reduction |
| K-Means2 | K-Means Clustering After Dimension Reduction |
| Hierarchical1 | Hierarchical Clustering Euclidean Before Dimension Reduction |
| Hierarchical2 | Hierarchical Clustering Euclidean After Dimension Reduction |
| Hierarchical3 | Hierarchical Clustering Agglomerative Complete Before Dimension Reduction |
| Hierarchical4 | Hierarchical Clustering Agglomerative Complete After Dimension Reduction |
| Hierarchical5 | Hierarchical Clustering Agglomerative Average Before Dimension Reduction |
| Hierarchical6 | Hierarchical Clustering Agglomerative Average After Dimension Reduction |
| Hierarchical7 | Hierarchical Clustering Agglomerative Ward Before Dimension Reduction |
| Hierarchical8 | Hierarchical Clustering Agglomerative Ward After Dimension Reduction |
| Gaussian1 | Gaussian Mixture Model Before Dimension Reduction |
| Gaussian2 | Gaussian Mixture Model After Dimension Reduction |
| Spectral1 | Spectral Clustering Radial basis function Before Dimension Reduction |
| Spectral2 | Spectral Clustering Radial basis function After Dimension Reduction |
| Spectral3 | Spectral Clustering Nearest Neighbor Before Dimension Reduction |
| Spectral4 | Spectral Clustering Nearest Neighbor function After Dimension Reduction |

In **Table 9**, all the legend descriptions for hyperparameter-tuned algorithm of **Fig. 12** and **Table 10** are given so that any reader can understand them easily, these are hyperparameter tuned.

Table 10. Result analysis of unsupervised algorithms

| Algorithms | Accuracy | Precision | Recall | F1 | Silhouette |
|---|---|---|---|---|---|
| K-Means1 | 0.4127 | 0.3036 | 0.3498 | 0.3251 | 0.3071 |
| K-Means2 | 0.5873 | 0.4922 | 0.6502 | 0.5603 | 0.4683 |
| Hierarchical1 | 0.3203 | 0.3296 | 0.6586 | 0.4393 | 0.4115 |
| Hierarchical2 | 0.315 | 0.3169 | 0.6003 | 0.4148 | 0.5661 |
| Hierarchical3 | 0.3275 | 0.3458 | 0.7438 | 0.4722 | 0.6297 |
| Hierarchical4 | 0.3386 | 0.3569 | 0.7926 | 0.4922 | 0.6677 |
| Hierarchical5 | 0.3583 | 0.3727 | 0.8587 | 0.5198 | 0.6781 |
| Hierarchical6 | 0.3586 | 0.374 | 0.87 | 0.5231 | 0.7005 |
| Hierarchical7 | 0.3203 | 0.3296 | 0.6586 | 0.4393 | 0.583 |
| Hierarchical8 | 0.315 | 0.3169 | 0.6003 | 0.4148 | 0.5661 |
| Gaussian1 | 0.6106 | 0.5325 | 0.3033 | 0.3864 | N/A |
| Gaussian2 | 0.6167 | 0.5433 | 0.3268 | 0.4081 | N/A |
| Spectral1 | 0.5648 | 0.4737 | 0.6878 | 0.561 | 0.0059 |
| Spectral2 | 0.4107 | 0.3779 | 0.4496 | 0.4782 | 0.0015 |
| Spectral3 | 0.4791 | 0.379 | 0.4512 | 0.4791 | 0.0092 |
| Spectral4 | 0.4782 | 0.3779 | 0.4496 | 0.4107 | 0.0083 |

From **Table 10**, it was observed that the Gaussian mixture algorithm outperformed all other algorithms in the case of accuracy which was **61%**. The Gaussian algorithm also achieved the highest precision score **54%**, where the precision score for K-means was **49%**. In the case of Recall, we got the highest score from the Hierarchical clustering algorithm, which was **86%**. Also, the K-means clustering algorithm conferred the best F1 score which was **56%**.

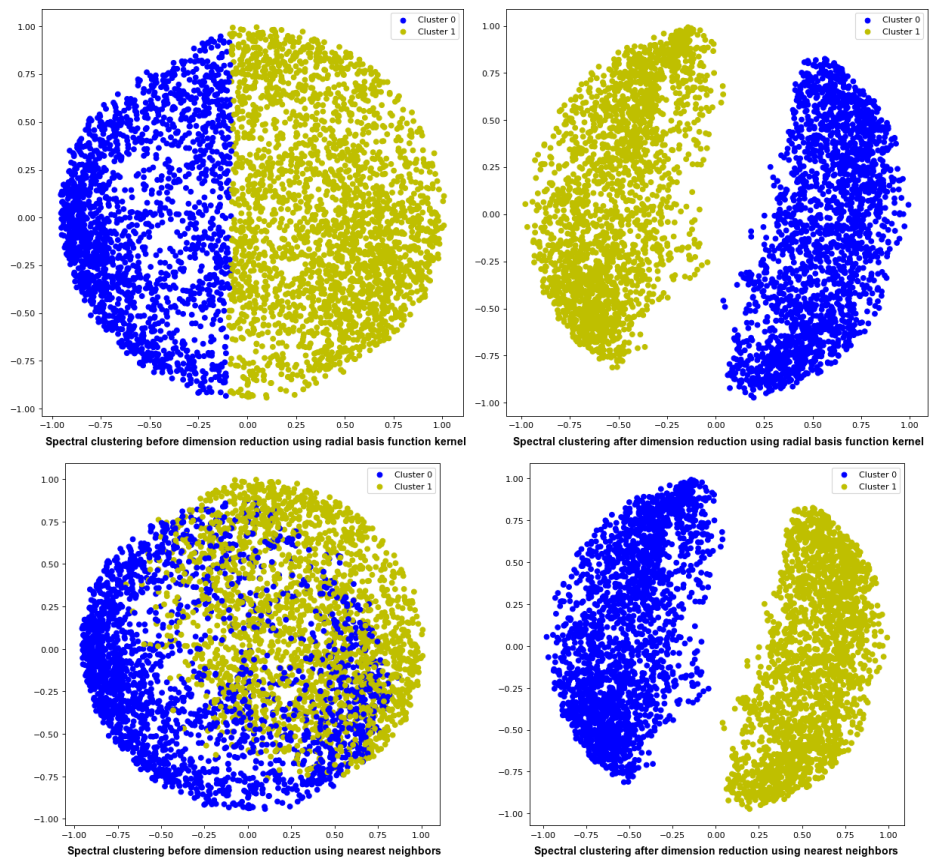

**Fig. 8.** Spectral clustering analysis by hyperparameter tuning. In **Fig. 8** we presented a visualization of spectral clustering analysis by hyperparameter tuning by showing scatter plots for applying each hyperparameter. Here you can see the changes in the outputs earlier and afterward of implementing the hyperparameter in spectral clustering. Two affinity matrices were used here for hyperparameter tuning, those are radial basis function kernel and nearest neighbors. The mentioned spectral clustering analysis was implemented on the basis of principal component analysis.

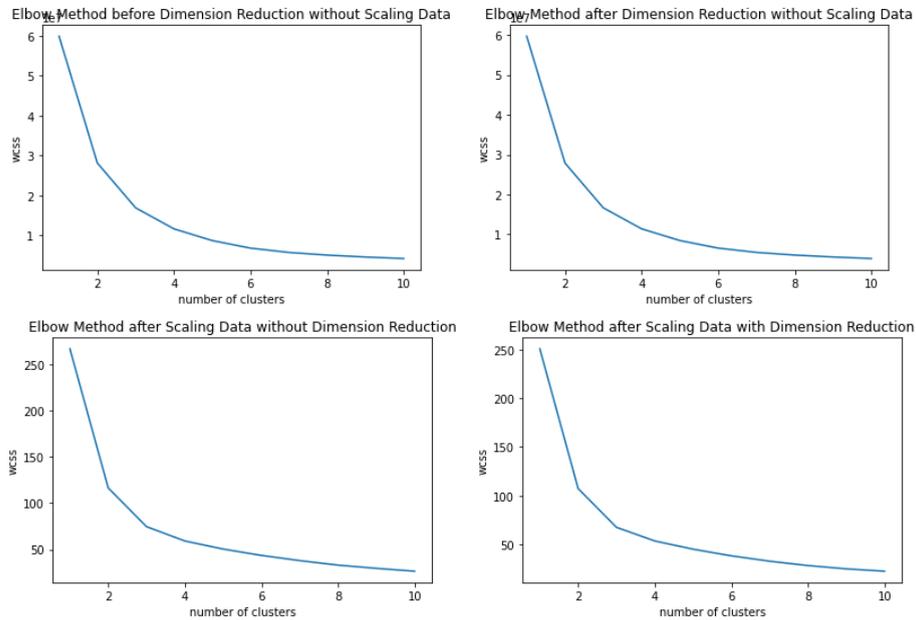

**Fig. 9.** Elbow method of K-means clustering analysis before and after data scaling and dimension reduction. In **Fig. 9** we presented a visualization of the Elbow method of K-means clustering analysis by using different parameters such as scaling data and dimension reduction. From **Fig. 9** it is observed that before scaling data and dimension reduction, the elbow was not sharp enough, although after scaling the data and reducing the dimensionality of the data, the elbow method worked better and those tunings helped to sharper the elbow which guided us to choose the perfect elbow for getting best results from K-means clustering analysis.

**Table 11.** Result Analysis of Supervised Classification Algorithms

| Algorithms | Accuracy | Precision | Recall | F1 |
|---|---|---|---|---|
| SVC | 0.72 | 0.72 | 0.52 | 0.60 |
| XGBoost | 0.74 | 0.72 | 0.59 | 0.65 |
| Naive Bayes | 0.69 | 0.66 | 0.51 | 0.57 |
| Light GBM | 0.75 | 0.74 | 0.61 | 0.67 |
| KNN | 0.73 | 0.76 | 0.46 | 0.58 |
| Decision Tree | 0.66 | 0.58 | 0.60 | 0.59 |
| CatBoost | 0.75 | 0.74 | 0.61 | 0.67 |
| AdaBoost | 0.66 | 0.58 | 0.61 | 0.59 |
| Logistic Regression | 0.72 | 0.71 | 0.52 | 0.61 |
| Generalized Linear Model | 0.69 | 0.78 | 0.68 | 0.72 |
| Random Forest | 0.76 | 0.75 | 0.59 | 0.66 |

From **Table 11**, it was observed that the Random Forest outperforms all other algorithms in the case of accuracy which was **76%**, the second-best was Light GBM here with a **75%** accuracy score. Moreover, the Generalized Linear algorithm achieves the highest precision score **78%**, where the second-best performance was shown by K-nearest Neighbor with a **76%** precision score. In the instance of the Recall score, the Generalized Linear Model provided the highest score of **68%**. Furthermore, this model had the highest F1 score of **72%**.

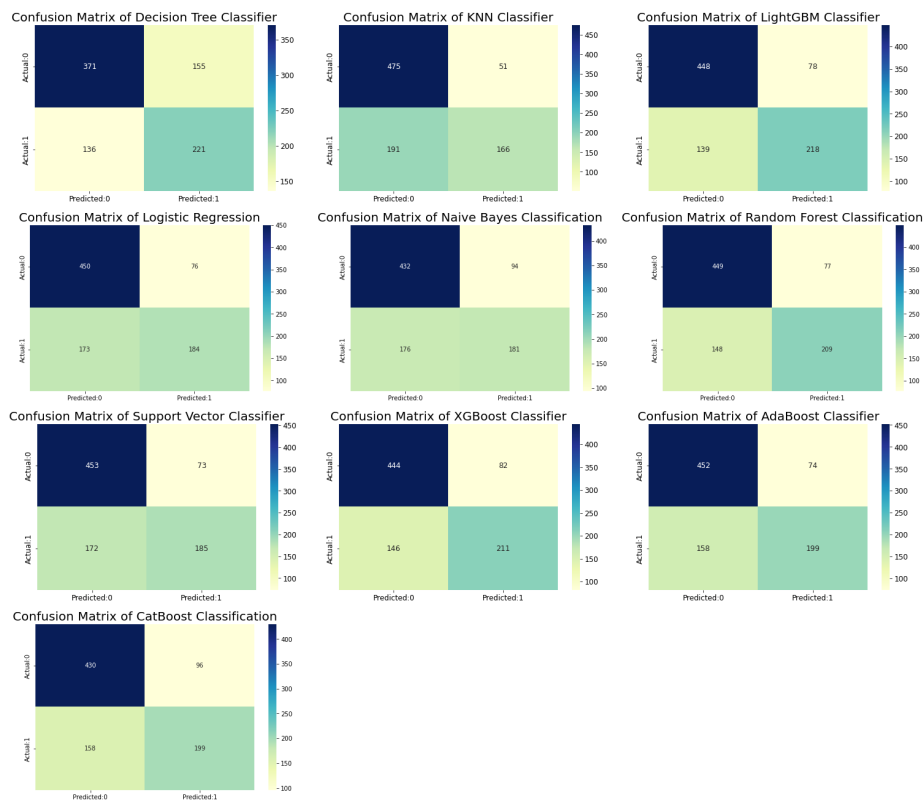

**Fig. 10.** Confusion matrix of all supervised classifiers. In **Fig. 10** we visualized the confusion matrix for each supervised classification that we used in this research. Observing all the confusion matrix for all classifiers, we came to the conclusion, that LightGBM was able to successfully predict the highest amount of correct predictions (**666** correct predictions out of **883** total predictions) and the second-highest performance was seen from Random Forest (**658** correct predictions out of **883** total predictions). The third highest performance was observed from XGBoost (**655** correct predictions out of **883** total predictions). In **Table 12** performance of all classifications according to the confusion matrix is visible in order of highest to lowest performance.

**Table 12.** Performance of classifiers according to confusion matrix.

| Classifiers | Correct Predictions | Total Predictions |
|---|---|---|
| LightGBM | 666 | 883 |
| Random Forest | 658 | 883 |
| XGBoost | 655 | 883 |
| AdaBoost | 651 | 883 |
| K-nearest Neighbors | 641 | 883 |
| Support Vector Machine | 638 | 883 |
| Logistic Regression | 634 | 883 |
| CatBoost | 629 | 883 |
| Naive Bayes | 613 | 883 |
| Decision Tree | 592 | 883 |

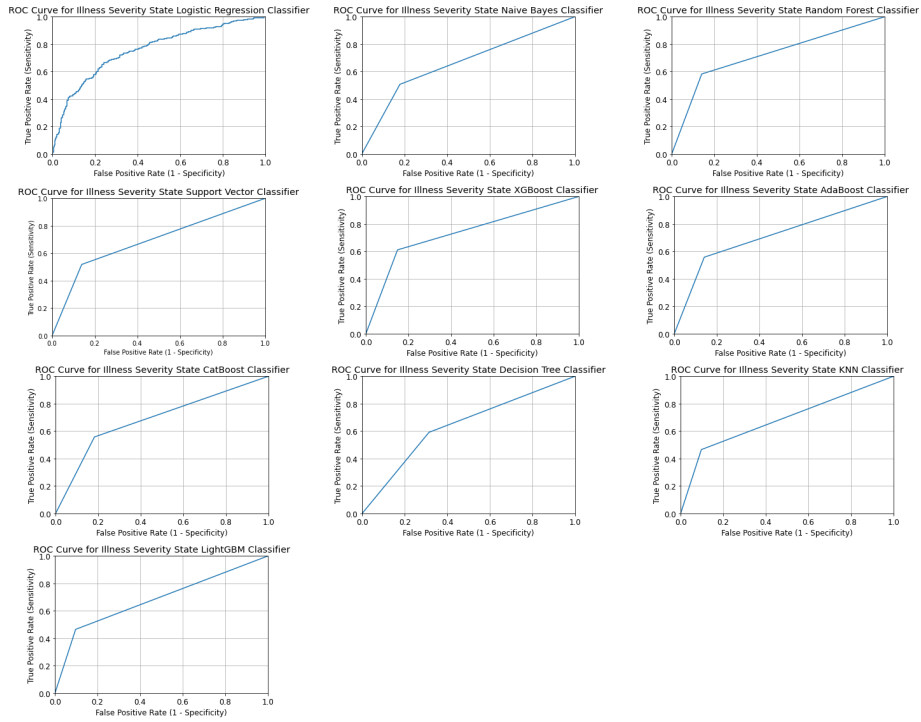

**Fig. 11.** ROC curve of all supervised classifiers. In **Fig. 11** we visualized the ROC curve for each supervised classification that we used in this research. Observing all ROC curves for all classifiers, we came to the conclusion that LightGBM and K-nearest Neighbors classifiers performed better than other classifiers.

**Table 13.** Result Analysis of Semi-Supervised Classification Algorithm

| Algorithms | Accuracy | Precision | Recall | F1 |
|---|---|---|---|---|
| Fast Large Margin | 0.72 | 0.72 | 0.84 | 0.78 |

From **Table 13**, we observed that compared to **Table 12** Fast Large Margin algorithm's Recall score **84%** and F1 score **78%** was far far better than any algorithm from **Table 12** though it was a semi-supervised approach, not fully supervised. Which was expected, as we all know already that semi-supervised approaches mostly performed better than any supervised or unsupervised methods. Semi-supervised models, as demonstrated by Al-Azzam N. and Shatnawi I., employ all of the data's readily available information to create the highest accurate prediction. Semi-supervised algorithms can obtain very highly accurate (**90–98%**) while just using half of the training examples [58].

**Table 14.** Result Analysis of Deep Learning Algorithm

| Algorithms | Accuracy | Precision | Recall | F1 |
|---|---|---|---|---|
| Multi-Layer Feed-Forward Artificial Neural Network | 0.75 | 0.75 | 0.87 | 0.81 |

From **Table 14**, it was clearly observable that Multilayer Feed-forward performed outstanding according to all performance scores having **75%** in the accuracy score, **75%** in the precision score, **87%** in recall score, **81%** in the F1 score.

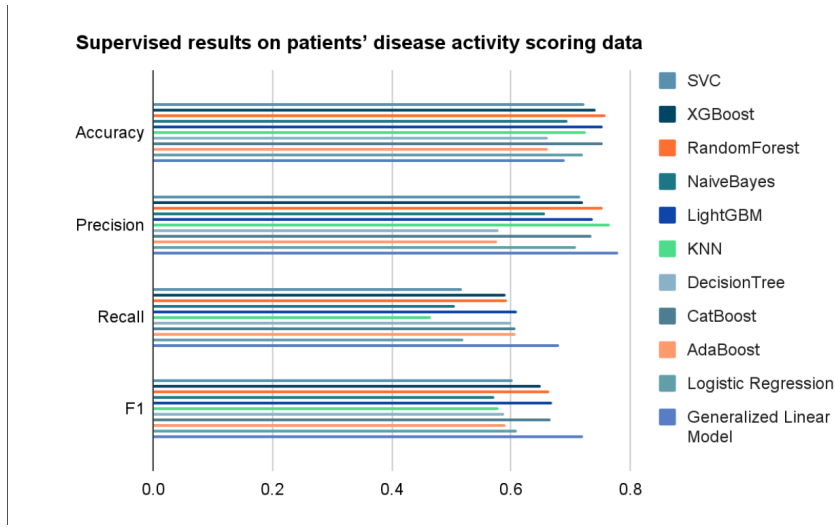

**Fig. 12.** Supervised results on patients' disease activity scoring data. In **Fig. 12** the visualization of the line chart shows that according to the accuracy score Random Forest performed the best, on the other hand, following precision, recall, and F1 score, Generalized Linear Model performed better than any approach.

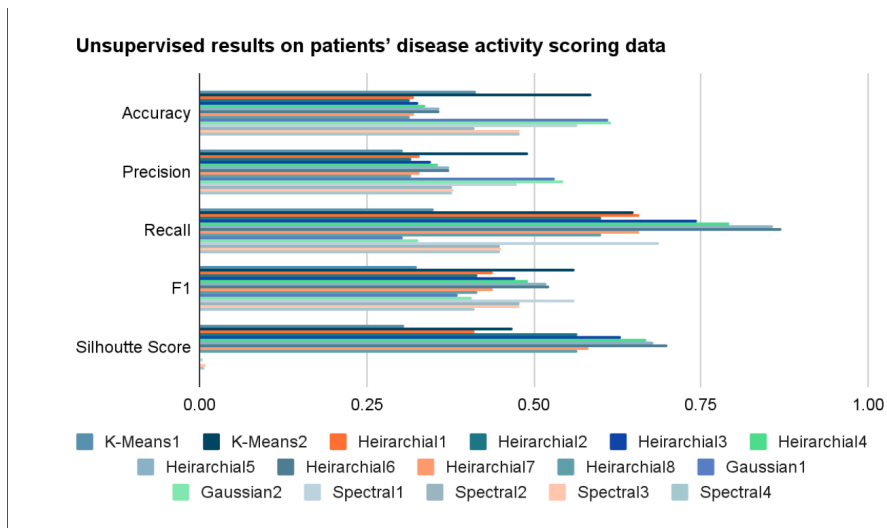

**Fig. 13.** Unsupervised results on patients' disease activity scoring data. In **Fig. 13** the visualization of the line chart shows that according to the accuracy score Gaussian2 (Gaussian Mixture Model After Dimension Reduction) performed the best, on the other hand, following precision score Higherarcical4 (Hierarchical Clustering Agglomerative Complete After Dimension Reduction) performed better than any approach in unsupervised. According to the recall score, Higherarcial6 (Hierarchical Clustering Agglomerative Average After Dimension Reduction) outperformed other unsupervised approaches. Additionally, following the F1 score, K-means2 (K-Means Clustering After Dimension Reduction) and Spectral1 (Spectral Clustering with Radial Basis Function Before Dimension Reduction) performed better than others. Lastly, for evaluating all the clustering techniques here, focusing on the Silhouette score, the more it is near to **+1**, the better it is as a clustering technique. On the other hand, the more it is near to **-1**, the worse the clustering technique it is and if the score is near zero, then the clusters are overlapping. By following the Silhouette score, we had observed that Higherarcial6 (Hierarchical Clustering Agglomerative Average After Dimension Reduction) secured the best position while Higherarchial5 (Hierarchical Clustering Agglomerative Average Before Dimension Reduction) secured the second-highest position in the ranking of clustering techniques.

## 5 Limitations and Future Work

In the future, we will try to collect more real-time datasets from other hospitals so that the models can be trained better to predict the categorization of illness severity states of patients. As a limitation of this research, we can include that we just worked on one dataset from one private medical company from Indonesia, which we can not rely on in case of providing automated medical services to patients as it can be risky. Although if we can manage more data from several medicals the data range should be larger. In this dataset we have **4, 412** patients' data, in the future we are planning to work with more than **10, 000** patients' data to come up with better outputs. Another limitation of this research work was we only used one method for implementing the semi-supervised machine learning technique and one method for implementing the deep learning technique. For future research, we want to use several deep learning and semi-supervised techniques. Additionally, we only trained our models, but could not deploy them into our IoMT devices successfully, though we tried to, as we had a shortage of time, we decided to deploy our custom models of classification illness severity in the future. Furthermore, the dataset should have more variety, as it was collected from another research work from Indonesian researchers [6], the dataset was from a private hospital and from a specific country if data can be collected from different countries and different types of clinics such as government, private, military clinics, this research would be more impactful. Because then the dataset will have more varieties of patients who belong to different classes of society. It would be more realistic research then.

## 6 Conclusion

This research proposes and effectively implements a performance analysis of categorizing the illness severity from their electronic health record data using supervised, unsupervised, semi-supervised machine learning as well as deep learning techniques. This study is planned and carried out with the help of clinical report data such as the patient's Hematocrit, Hemoglobin, Erythrocyte, Thrombocytes, and all other measurements of different cells in the blood. Our proposed approach can analyze and forecast the severity of patients' illness in situations when there is a scarcity of medical services because of the large number of patients and medical staff cannot give all of the patient's services in the hospital because of limited spaces, therefore we must make quick decisions about who requires emergency care and who does not. In this case, rather than relying on the human brain decisions of medical staff, this method can expedite things as an automated procedure. Patients with less severity of the illness can simply be informed of their condition and taken home; on the other hand, when patients are at high risk, they will get higher priority to get medical service first. From this research, we found multiple supervised, unsupervised, semi-supervised machine learning and deep learning algorithms are useful for categorizing patients' illness severity states as a result of our research. In a short summary from result analysis, from unsupervised clustering techniques, hyperparameter-tuned Hierarchical with 86% precision score and Gaussian Mixture Model with 61% accuracy worked really well. Noticeably, dimension reduction helped a lot to improve results, especially in the case of unsupervised methods. Furthermore, from supervised learning techniques, hyperparameter-tuned Random Forest outperformed all other supervised techniques at the accuracy of 0.76, also according to precision score Generalized Linear algorithm performed the best with 78% score. The Semi-supervised approach Fast Large Margin performed unexpectedly well with 84% recall score and 78% F1 score. Lastly, the Deep learning approach Multi-layer Feed-forward Neural Network performed outstandingly with 75% accuracy, 75% precision, 87% recall, 81% F1 score. As our goal of this research was to improve the previous work [6] and try out different new models, we were able to do both and from our observations, we found out in previous work hypermeter-tuned XGBoost was performing best, but in our experiments, Random Forest performed best among supervised classifiers with 76% accuracy. Also, in previous work [6], they only used supervised classification, but we tried out, supervised, unsupervised, semi-supervised machine learning models along with deep learning techniques, and from these new model experiments we got new insights which we have explained in detail already.